\documentclass[a4paper,journal]{IEEEtran}
\usepackage{amsmath,amsfonts}
\usepackage{algorithmic}
\usepackage{algorithm}
\usepackage{array}
\usepackage[caption=false,font=normalsize,labelfont=sf,textfont=sf]{subfig}
\usepackage{textcomp}
\usepackage{stfloats}
\usepackage{url}
\usepackage{verbatim}
\usepackage{graphicx}
\usepackage{cite}
\usepackage{fancyhdr}
\usepackage{ifthen}
\begin{document}

\title{Soft Electroadhesive Feet for Micro Aerial Robots Perching on Smooth and Curved Surfaces}

\author{Chen Liu~\IEEEmembership{Member, IEEE}, Sonu Feroz, Ketao Zhang~\IEEEmembership{Member, IEEE}
\thanks{This work is funded by the Innovate UK Project under grant No.10074371.\textit{(Corresponding author: Ketao Zhang.)}}
\thanks{Chen Liu, Sonu Feroz, and Ketao Zhang are with the Centre for Advanced Robotics at Queen Mary University of London, UK {\tt\small {chen.liu@qmul.ac.uk, s.feroz@se22.qmul.ac.uk, ketao.zhang}@qmul.ac.uk}}
}



\maketitle

\thispagestyle{fancy}

\begin{abstract}
Electroadhesion (EA) provides electrically switchable adhesion and is a promising mechanism for perching micro aerial robots on smooth surfaces. However, practical implementations of soft and stretchable EA pads for aerial perching remain limited. This work presents (i) an efficient workflow for fabricating soft, stretchable electroadhesive pads with sinusoidal-wave and concentric-circle electrodes in multiple sizes, (ii) a controlled experimental comparison of normal and shear adhesion under inactive (0~kV) and active (4.8~kV) conditions using an Instron-based setup, and (iii) a perching demonstration using a Crazyflie quadrotor equipped with electroadhesive feet on flat and curved substrates. Experimental results show that shear adhesion dominates, reaching forces on the order of 3~N with partial pad contact, while normal adhesion is comparatively small and strongly dependent on substrate properties. The Crazyflie prototype demonstrates repeatable attachment on smooth plastic surfaces, including curved geometries, as well as rapid detachment when the voltage is removed. These results highlight the potential of soft electroadhesive feet for lightweight and reliable perching in micro aerial vehicles (MAVs).
\end{abstract}


\section{Introduction}

\IEEEPARstart{P}{erching} is an important ability for an aerial robot to temporarily attach to surrounding structures, which can extend mission duration by reducing the need for continuous hovering and enabling energy-efficient observation or sensing tasks~\cite{wuest2024agile}. This capability is particularly valuable for micro aerial vehicles (MAVs), which are strongly constrained by battery capacity and payload~\cite{shen2024sunlight,wang2024novel,zhang2022aerial}. By landing and attaching to nearby structures, MAVs can conserve energy while maintaining situational awareness during long-duration missions~\cite{hang2019perching,zhang2019bioinspired}.

\begin{figure}[t]
\centering
\includegraphics[width= 1\columnwidth]{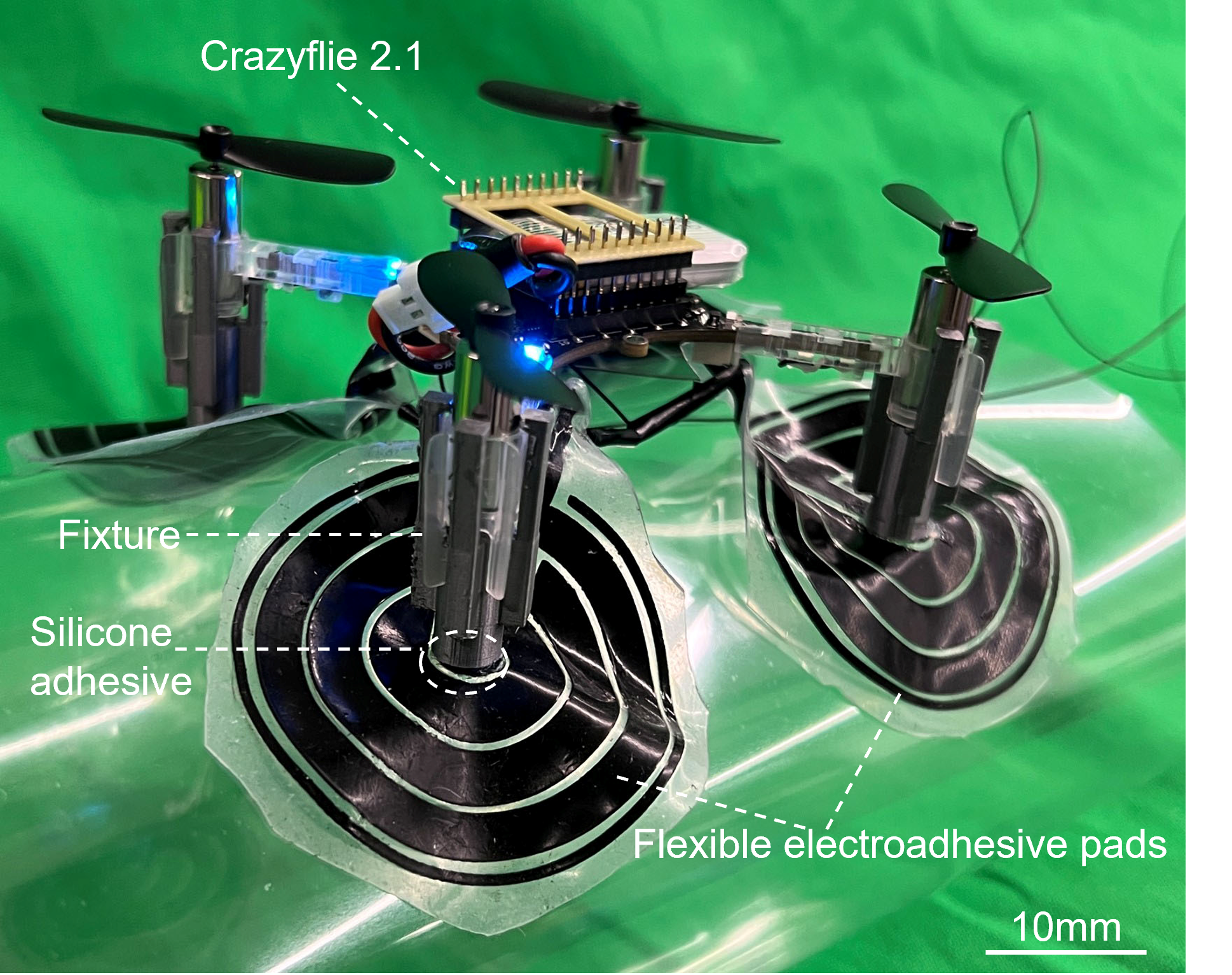}
\caption{Crazyflie quadrotor equipped with soft electroadhesive pads demonstrating conformal attachment to a curved plastic surface.} 
\label{fig1}
\end{figure}

Despite extensive research on aerial robot perching, achieving reliable attachment in real environments remains challenging. Many existing perching mechanisms rely on mechanical gripping strategies such as claws, spines, or micro-hooks that latch onto surface irregularities. While these approaches can perform well on rough or textured substrates, they are far less effective on smooth planar surfaces such as glass, plastic panels, or coated metal sheets~\cite{bai2024design,speciale2025review,lynch2024powerline}. These surfaces are common in built environments but provide little mechanical anchoring, making stable perching difficult~\cite{luo2025high}. In addition, many mechanical perching mechanisms introduce additional weight, mechanical complexity, or require specific surface properties, limiting their applicability for lightweight MAV platforms~\cite{askari2023avian}.

Electroadhesion (EA) offers a promising alternative for controllable surface attachment. By applying a high voltage across patterned electrodes embedded in a dielectric layer, an electric field is generated that induces attractive forces between the pad and the target surface. Unlike mechanical gripping mechanisms, EA can operate effectively on smooth surfaces and allows attachment and detachment to be controlled electrically without moving parts~\cite{liu20213d,caruso2025electroadhesion,liu2022electro}. These characteristics make electroadhesion particularly attractive for perching applications in micro aerial robots~\cite{guo2019electroadhesion}.

However, practical implementations of electroadhesion for MAV perching remain limited. Many EA pads reported in the literature are relatively rigid, which can reduce the effective contact area when interacting with curved or uneven surfaces. In addition, fabrication methods for soft and stretchable EA pads are often time-consuming and difficult to scale, making it challenging to systematically explore different electrode geometries and pad sizes~\cite{park2020lightweight,liu2023electric,graule2016perching,liu2018novel}.

In this work, we investigate soft and stretchable electroadhesive pads designed for MAV perching on smooth and curved surfaces, as shown in Fig.~\ref{fig1}. Two electrode wave (Fig.~\ref{fig2}(b)) and concentric circular (Fig.~\ref{fig2}(c)) patterns are fabricated in multiple sizes using an efficient fabrication workflow. The pads are experimentally characterised using an Instron-based setup to measure both normal and shear adhesion under inactive (0 kV) and active (4.8~kV) conditions. The results show that shear adhesion dominates, reaching forces on the order of 3~N with partial pad contact, while normal adhesion is comparatively small and strongly dependent on substrate properties.

To improve perching reliability, the proposed system integrates four electroadhesive feet on the landing legs of a Crazyflie 2.1 (Bitcraze, Sweden) quadrotor. This distributed configuration increases the probability of successful attachment by providing multiple contact opportunities during landing. Even if some pads fail to establish full adhesion, attachment by one or more remaining pads can still stabilise the vehicle, thereby introducing redundancy and improving tolerance to partial adhesion failure.

The main contributions of this work are:

\begin{enumerate}

\item[1)] Soft electroadhesive pad design: Development of stretchable EA pads with sinusoidal-wave and concentric-circle electrode geometries.
\item[2)] Experimental characterisation: Controlled measurement of normal and shear adhesion under 0 kV and 4.8 kV conditions using an Instron testing platform. 
\item[3)] Reliable MAV perching demonstration: Integration of four EA feet on a Crazyflie quadrotor, demonstrating attachment on smooth and curved surfaces with improved reliability through distributed contact.
\end{enumerate}

\section{System Concept and EA Pads Design}
\subsection{Pad Structure and Materials}
\begin{figure}
\centering
\includegraphics[width= 0.9\columnwidth]{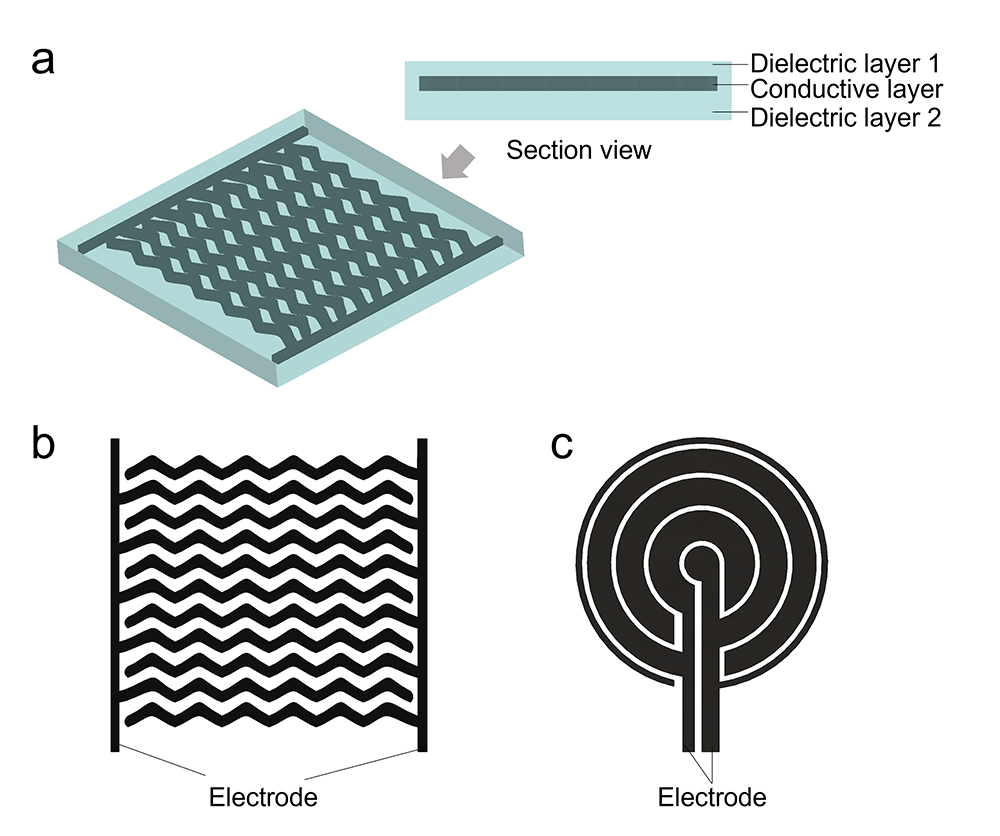}
\caption{Soft electroadhesive (EA) pad design: (a) multilayer structure and section view; (b) wave electrode pattern; (c) concentric circular electrode pattern.}
\label{fig2}
\end{figure}

The soft electroadhesive (EA) pad is designed as a multilayer structure consisting of two dielectric layers and an embedded conductive electrode layer, as illustrated in Fig.~\ref{fig2}(a). The conductive layer forms the patterned electrodes, which are connected to a high-voltage power supply to generate electroadhesion. The electrodes are encapsulated between two dielectric layers, providing electrical insulation and mechanical compliance. The bottom dielectric layer (dielectric layer 2) defines the distance between the electrodes and the surrounding environment, while the top dielectric layer (dielectric layer 1) plays a critical role in determining the effective gap between the electrodes and the target surface. In particular, the thickness of the top dielectric layer directly influences the separation between the electrode and the object surface, which strongly affects the generated electroadhesive force.

By adopting a soft and compliant dielectric material, the pad can deform to match surface irregularities and curved geometries, thereby improving the effective contact area during attachment.

\subsection{Electrode Pattern Design}
The geometry of the electrodes significantly influences the distribution of electric fields and the resulting adhesion performance. In this work, two electrode patterns are investigated: a wave pattern (Fig.~\ref{fig2}(b)) and a concentric circular electrode pattern (Fig.~\ref{fig2}(c)).

The wave pattern provides continuous electrode paths with smooth transitions, enabling flexibility and good stretchability in the soft structure. The absence of sharp corners helps reduce localized electric field concentration~\cite{minaminosono2022fabrication}. The concentric-circle pattern offers radial symmetry and eliminates sharp edges, resulting in a more uniform distribution of fringe electric fields. Compared with geometries containing corners, this design reduces the risk of dielectric breakdown and improves the consistency of adhesion forces across the contact area~\cite{ruffatto2013optimization}.

In addition, the circular geometry is more suitable for aerial robot perching, as it promotes uniform contact and reduces edge lifting when the pad is partially attached to the surface.

\subsection{Electroadhesion Principle}
\begin{figure}
\centering
\includegraphics[width= 1\columnwidth]{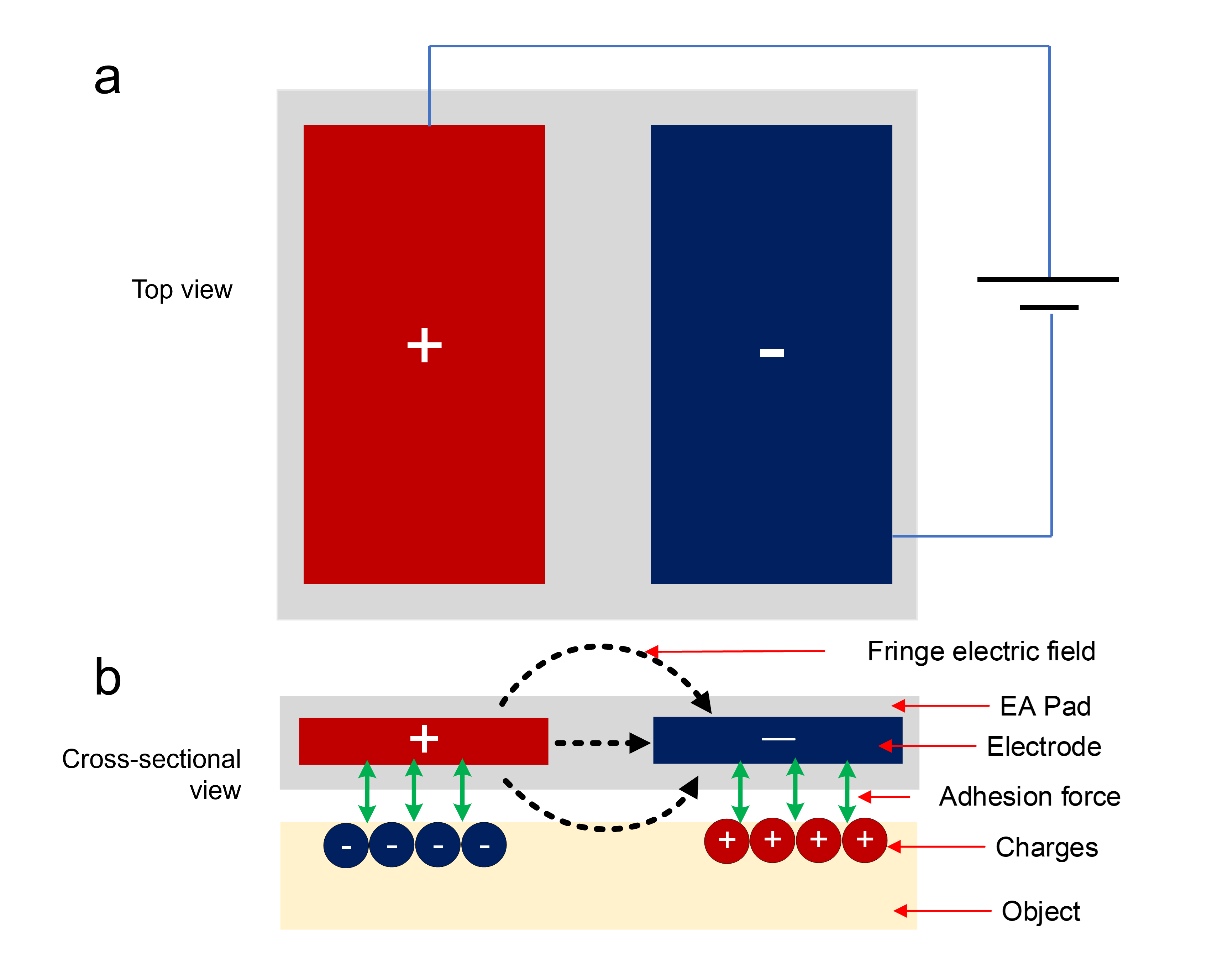}
\caption{Working principle of the electroadhesive (EA) pad: (a) top view of the electrode layout and (b) cross-sectional view. The applied high voltage generates fringe electric fields that induce charge polarization on the substrate surface, resulting in electroadhesive forces.}
\label{fig3}
\end{figure}

Electroadhesion is achieved through electrostatic interactions between the electrodes and the target surface. When a high voltage is applied across electrodes of opposite polarity, fringe electric fields are generated near the electrode edges, as illustrated in Fig.~\ref{fig3}. These electric fields polarize the surface of the object and induce opposite charges. The electrostatic attraction between the induced charges and the electrodes results in an adhesion force that holds the pad against the surface. The electroadhesive force can be approximated using analytical models derived from electrostatic theory~\cite{guo2019electroadhesion}:

\begin{equation}
F=\frac{1}{2}n\varepsilon_{0}l[(\frac{\varepsilon_r}{\varepsilon_0})^2-1]\overline{C}({\frac{w}{w+s}},\frac{2(t+d_t)}{w+s})(E^{air}_{BD})^2
\end{equation}
where $n$ denotes the number of EA periods,
$l$ is the substrate width,
$w$ is the electrode width,
$s$ is the electrode gap,
$t$ is the dielectric thickness,
$d_t$ represents the air gap between the dielectric and the substrate,
$E^{\mathrm{air}}_{BD}$ is the air breakdown electric field strength,
$\varepsilon_0$ is the vacuum permittivity,
and $\varepsilon_r$ is the relative permittivity of the dielectric.

This relationship indicates that the electroadhesive force depends on the electric field strength, dielectric properties, and electrode geometry. In particular, reducing the gap between the electrode and the object surface and increasing the effective electrode area can significantly enhance adhesion performance.

\subsection{Multi-Feet Perching Strategy}
To improve the robustness of aerial robot perching, four electroadhesive pads are integrated into the landing legs of a Crazyflie quadrotor, forming electroadhesive feet (Fig.~\ref{fig1}). Unlike single-pad designs, the proposed configuration introduces redundancy in the attachment process. During landing, not all pads are guaranteed to achieve full contact due to surface irregularities, misalignment, or dynamic effects. However, the presence of multiple pads ensures that attachment can still be achieved even if only one or a subset of the pads establishes effective contact.

This distributed adhesion mechanism significantly increases the probability of successful perching and improves tolerance to partial adhesion failure. Additionally, multiple contact points provide additional stability by distributing adhesion forces across the structure, enabling reliable attachment on both flat and curved surfaces.

\section{Fabrication and Materials}

\begin{figure*}[t]
\centering
\includegraphics[width= 2\columnwidth]{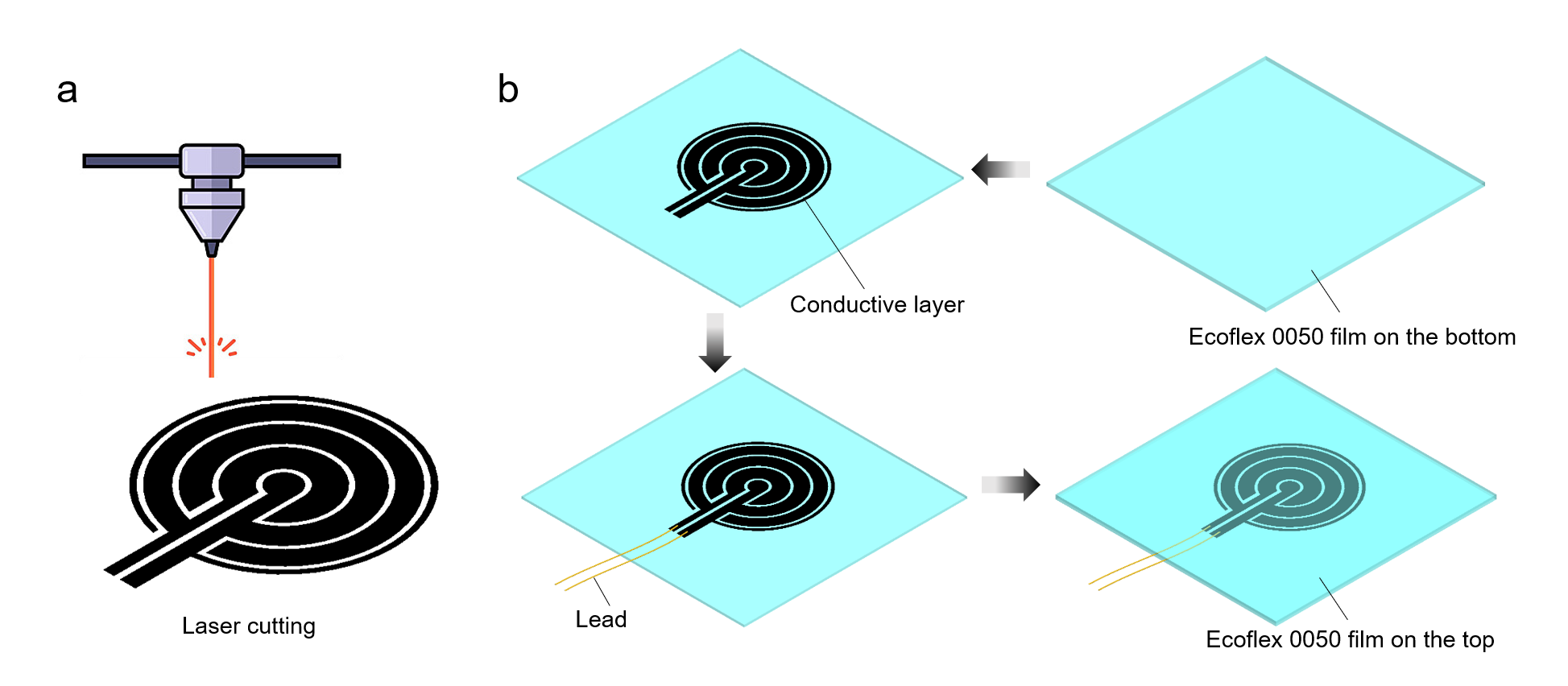}
\caption{Fabrication process of the soft electroadhesive pad: (a) laser cutting of the conductive silicone electrode pattern; (b) casting of the laminated structure by placing the patterned electrode on the bottom Ecoflex layer, attaching the lead, and encapsulating it with a top Ecoflex layer.} 
\label{fig4}
\end{figure*}

The soft electroadhesive (EA) pads were fabricated by combining laser cutting and film casting, as shown in Fig.~\ref{fig4}. Each pad consists of a patterned conductive electrode embedded between two Ecoflex dielectric layers. A conductive silicone sheet (0.3~mm thick, TYM, UK) was used as the electrode material, while Ecoflex 00-50 (Smooth-On, USA) was used for both the bottom and top dielectric layers. The electrode patterns were first designed in SolidWorks and cut from the conductive silicone sheet using a laser cutter. Two electrode geometries were used in this work: a wave pattern and a concentric-circular pattern. For the wave electrode, an additional laser-cut stencil was prepared to maintain the spacing between adjacent conductive tracks and to assist in positioning during assembly. For the concentric-circular electrode, temporary supporting lines were included in the cutting design to improve handling stability and were removed manually after placement of the electrode.

Ecoflex 00-50 Parts A and B were mixed at a 1:1 ratio and degassed under vacuum before casting. A bottom Ecoflex layer with a thickness of approximately 0.2~mm was cast onto the substrate using an Elcometer 4340 Automatic Film Applicator at a coating speed of 1~m/min, followed by curing at 40~$^\circ$C for 45~min. After curing, the patterned conductive electrode was placed onto the bottom dielectric layer and connected to electrical leads. A second Ecoflex layer, also approximately 0.2~mm thick, was then cast on top to fully encapsulate the electrode, followed by a second curing step at 40~$^\circ$C for 45~min. After curing, the completed EA pads were peeled from the substrate. The final devices had a total thickness of approximately 0.7~mm, including the 0.3~mm conductive silicone layer and the two Ecoflex layers. This method provided a simple and repeatable way to fabricate thin, compliant, and fully encapsulated EA pads with different electrode geometries for attachment to micro aerial robots.

\section{Experiments}

\subsection{EA pads test}

\begin{figure}
\centering
\includegraphics[width= 0.85\columnwidth]{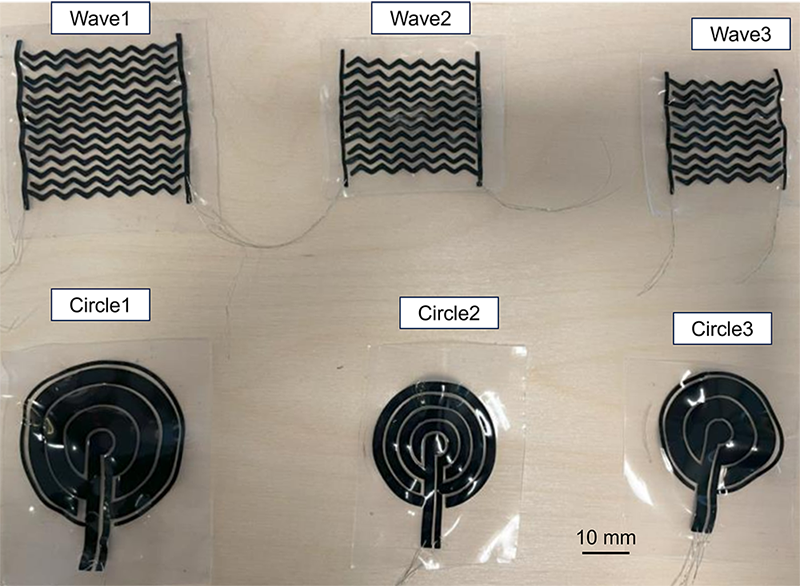}
\caption{Six fabricated soft electroadhesive (EA) pads used for pad-level characterisation, including three wave-pattern designs (Wave~1--Wave~3) and three concentric-circular designs (Circle~1--Circle~3) with different electrode geometries and sizes.}
\label{fig5}
\end{figure}

Before system-level integration, six fabricated EA pads were characterised to examine the influence of electrode geometry and pad size on electroadhesion performance, as shown in Fig.~\ref{fig5}. The tested samples included three wave-pattern pads (Wave~1--Wave~3) and three concentric-circular pads (Circle~1--Circle~3). Although a larger electrode area is generally expected to improve adhesion, the pad dimensions in this work were selected to account not only for area variation but also for fabrication feasibility and the limited mounting space on the Crazyflie platform. For the wave-pattern group, the electrode areas were 1300~mm$^2$ for Wave~1, 907~mm$^2$ for Wave~2, and 703~mm$^2$ for Wave~3. For the concentric-circular group, Circle~1 was the largest design, with varying electrode widths and an overall electrode area of approximately 1520~mm$^2$, while Circle~2 and Circle~3 had electrode areas of approximately 965~mm$^2$ and 968~mm$^2$, respectively. Circle~2 was designed with a more uniform electrode-width distribution, whereas Circle~3 retained the same general circular layout as Circle~1 but with a reduced outer structure. In this way, the six samples allowed the effects of both electrode pattern and effective electrode area to be examined systematically.

All pad-level experiments were carried out using an Instron 3342 testing machine. An RS Pro DC programmable high-voltage power supply was used to energise the pads, and each sample was tested under both inactive (0~kV) and active (4.8~kV) conditions. The voltage of 4.8~kV was selected because it produced clear electroadhesion during preliminary trials without causing dielectric breakdown. A smooth plastic substrate was used as the main contact surface, since preliminary comparison showed stronger and more repeatable adhesion on plastic than on wood. The rougher wooden surface led to noticeably weaker attachment.

\begin{figure}
\centering
\includegraphics[width= 0.85\columnwidth]{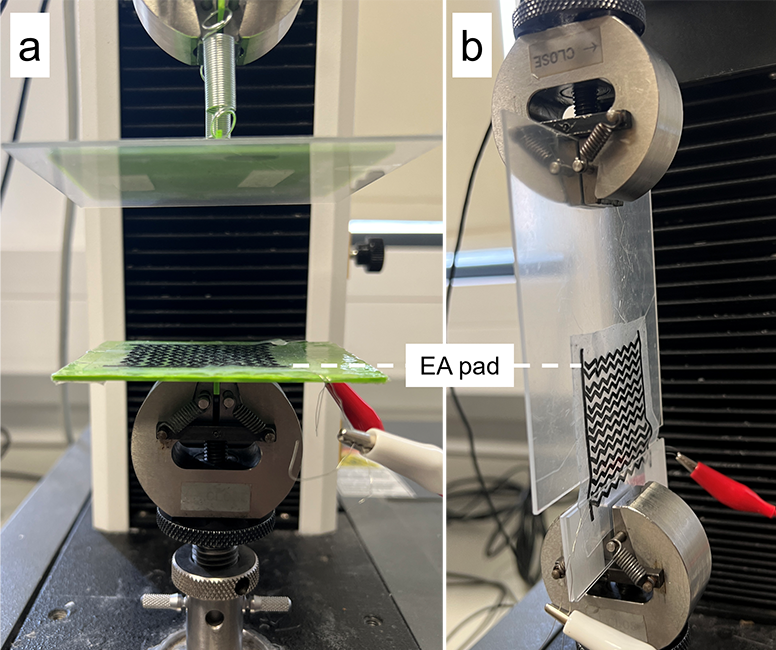}
\caption{Experimental setups for pad-level characterisation: normal-force test and shear-force test using the Instron platform.}
\label{fig6}
\end{figure}

Two types of force measurement were performed, namely normal-force testing and shear-force testing, as shown in Fig.~\ref{fig6}. In the normal-force test, the EA pad was fixed on the stationary side of the Instron, while the plastic substrate was mounted on the moving side (Fig.~\ref{fig6}(a)). The upper stage was then driven upward until detachment occurred, and the peak force was recorded. In the shear-force test, the substrate was mounted vertically and the pad was arranged in a corresponding vertical configuration (Fig.~\ref{fig6}(b)). To maintain alignment during loading, the base of the pad was supported between two microscope slides.
Each pad was tested first at 0~kV and then at 4.8~kV so that the passive adhesion of the soft structure could be separated from the additional adhesion generated by the applied electric field. In the shear-force test, full-area contact sometimes produced forces above 6~N and increased the risk of damaging the pads. Therefore, only approximately half of the pad area was kept in contact with the substrate during shear testing. Each condition was repeated five times in both normal and shear modes. The pad surfaces were not cleaned between repeated tests, so the results also reflected the effect of the practical pad condition during repeated use.
For the wave-pattern pads, the average normal force at 0~kV was 0.320~N for Wave~1, 0.309~N for Wave~2, and 0.255~N for Wave~3. Under 4.8~kV, these values increased slightly to 0.332~N, 0.372~N, and 0.278~N, respectively. The voltage-induced increase in normal force was therefore limited for all three wave designs. A much clearer voltage effect was observed in the shear-force measurements. At 0~kV, the average shear force was 1.221~N for Wave~1, 0.695~N for Wave~2, and 1.655~N for Wave~3. Under 4.8~kV, the shear force increased to 2.397~N for Wave~1, 3.314~N for Wave~2, and 2.250~N for Wave~3. Among the wave-pattern pads, Wave~2 gave the highest average shear force, indicating that the intermediate-sized wave design offered the best overall balance in this group.
For the concentric-circular pads, the average normal force at 0~kV was 0.308~N for Circle~1, 0.273~N for Circle~2, and 0.281~N for Circle~3. At 4.8~kV, the corresponding values became 0.324~N, 0.288~N, and 0.283~N. As in the wave-pattern group, only a small increase in normal force was observed under voltage. In contrast, the shear-force results again showed a much stronger voltage dependence. At 0~kV, the average shear force was 1.300~N for Circle~1, 0.354~N for Circle~2, and 0.374~N for Circle~3. Under 4.8~kV, these values increased to 3.045~N for Circle~1, 1.237~N for Circle~2, and 1.880~N for Circle~3. Circle~1 therefore showed the strongest shear performance within the circular group, while Circle~2 gave the weakest overall result.

\begin{figure}
\centering
\includegraphics[width=0.95\columnwidth]{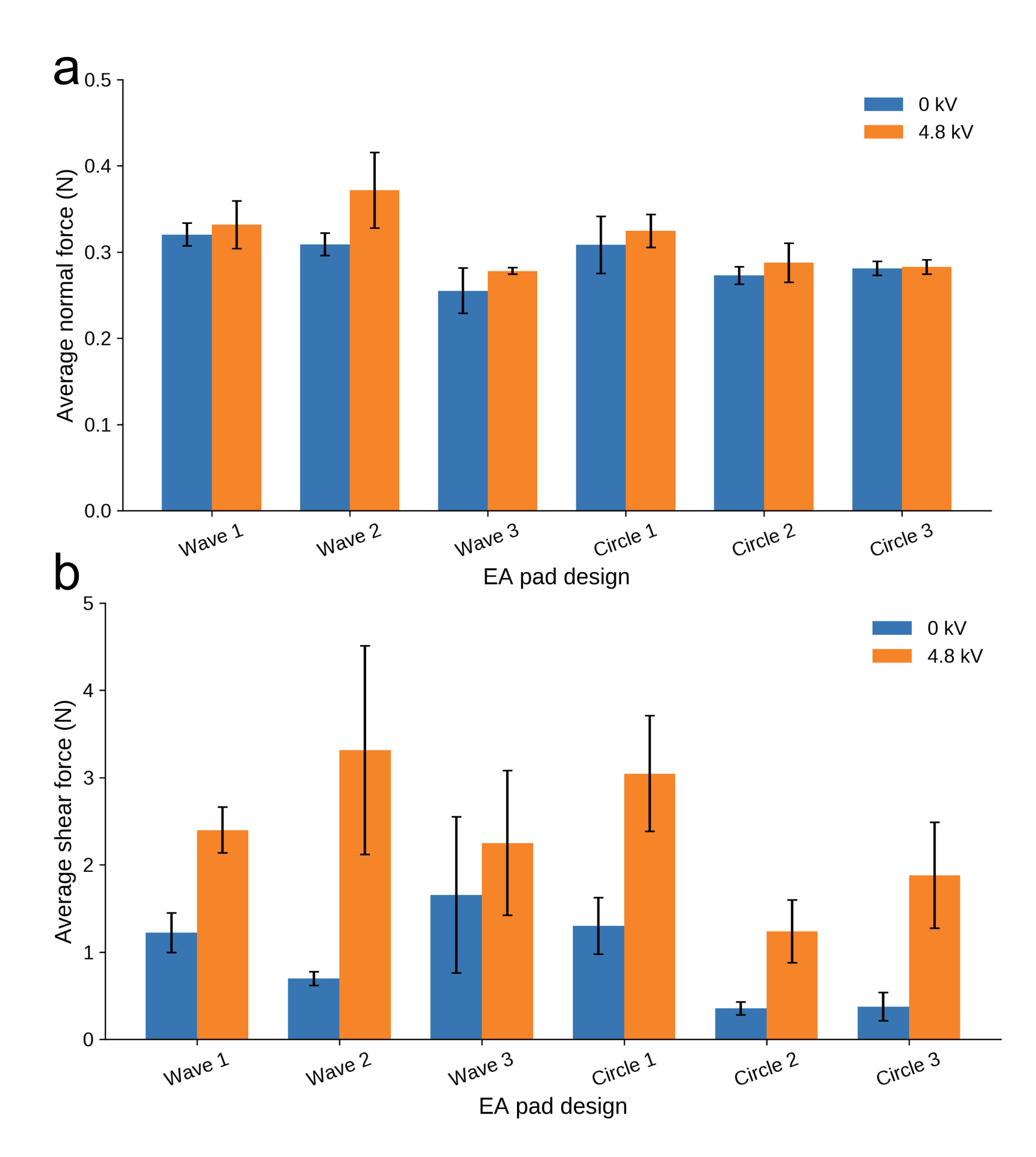}
\caption{Comparison of the average (a) normal force and (b) shear force generated by the six fabricated EA pads under 0~kV and 4.8~kV, respectively. Error bars indicate the standard deviation from repeated tests.}
\label{fig7}
\end{figure}

As summarised in Fig.~\ref{fig7}, the shear force was consistently much higher than the normal force for all six pads, indicating that the attachment behaviour of the soft EA pads was dominated by shear adhesion. Within the wave-pattern group, Wave~2 produced the highest shear force. Within the concentric-circular group, Circle~1 showed the best overall performance, combining strong shear adhesion with relatively stable behaviour during repeated tests. Although Wave~2 achieved the highest average shear force among all samples, Circle~1 was selected for the subsequent system validation experiments because its circular geometry was more suitable for integration as an electroadhesive foot on the landing leg of the Crazyflie, while still providing strong adhesion performance.

\subsection{System validation test}

\begin{figure}
\centering
\includegraphics[width=1\columnwidth]{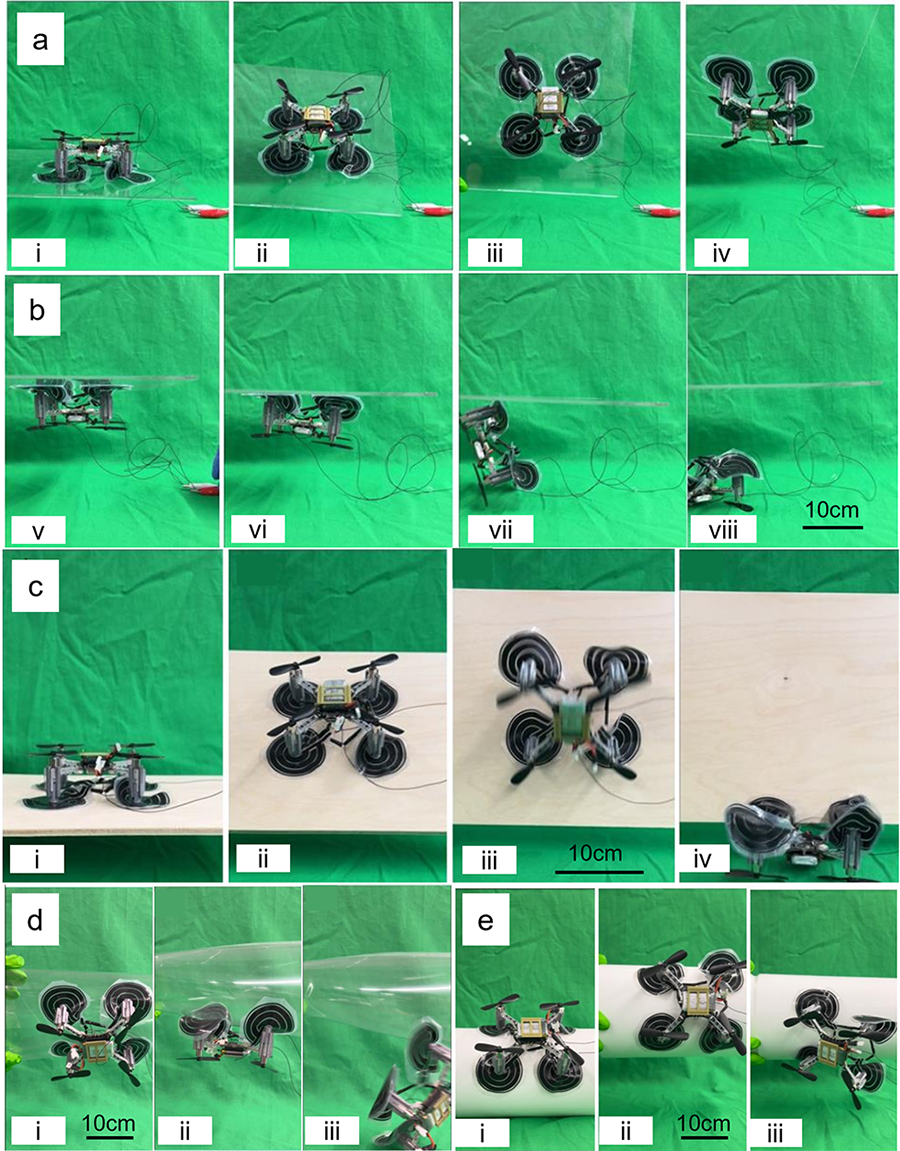}
\caption{System-level validation of the Crazyflie equipped with four soft electroadhesive feet based on the Circle~1 design. (a) Attachment on a flat plastic surface as the substrate was rotated from the horizontal to the vertical and fully inverted configurations. (b) Detachment sequence after the applied voltage was removed. (c) Attachment test on a flat wooden surface, showing weaker support and eventual detachment. (d) Attachment on a rolled plastic surface, including side views and a close-up showing conformal contact between the soft pads and the curved substrate. (e) Attachment test on a cylindrical PVC bottle, where vertical attachment was achieved but stable support in more demanding orientations was not maintained.}
\label{fig8}
\end{figure}

Based on the pad-level characterisation results, Circle~1 was selected for system-level validation. Although Wave~2 produced the highest average shear force among all tested samples, Circle~1 gave the best overall performance within the concentric-circular group and showed more stable behaviour during repeated tests. In addition, its compact and symmetric geometry was more suitable for integration on the landing legs of the Crazyflie. It provided a more uniform contact region at the foot and reduced the tendency for local edge lifting, which was especially important when attaching to curved substrates.

For the system demonstration, four Circle~1 pads were mounted onto the landing legs of a Crazyflie 2.1 quadrotor using 3D-printed fixtures, as shown in Fig.~\ref{fig8}. The pads were bonded to the fixtures using silicone adhesive, and the positive and negative terminals of the four pads were connected in parallel to an external high-voltage supply (EMCO A50, XP Power, UK). The total mass of the assembled system was approximately 43~g, corresponding to a gravitational load of about 0.42~N.

The first validation test was carried out on a flat plastic surface, which had already shown favourable adhesion in the pad-level experiments. As shown in Fig.~\ref{fig8}(a), the Crazyflie equipped with four EA feet remained attached while the plastic substrate was rotated from the horizontal position to the vertical and fully inverted configurations. The detachment behaviour after voltage removal is shown in Fig.~\ref{fig8}(b). Once the applied voltage was switched off, the robot could no longer remain attached and fell from the surface. This confirms that the observed attachment was generated by active electroadhesion rather than passive friction alone.
The same procedure was then repeated on a flat wooden surface. As shown in Fig.~\ref{fig8}(c), 
the attachment on wood was much weaker than that on plastic, and stable support could not be maintained as the surface orientation became more demanding. This observation is consistent with the earlier pad-level measurements, which showed only limited force enhancement on wood under voltage. The lower performance on wood is mainly attributed to its rougher surface, which reduced the effective contact area of the soft EA pads.
The adaptability of the proposed EA feet to curved surfaces was further examined using a rolled plastic sheet and a cylindrical PVC bottle. On the rolled plastic surface, shown in Fig.~\ref{fig8}(d), the soft pads conformed well to the cylindrical geometry and were able to support the Crazyflie under curved-surface contact. The close-up view clearly shows that the compliant multilayer pads could maintain intimate contact with the rolled plastic surface. On the PVC bottle, shown in Fig.~\ref{fig8}(e), the system could achieve attachment on the curved cylindrical surface, but the stability was lower than that on the rolled plastic sheet and reliable support could not be maintained under more challenging orientations.
Overall, the system-level experiments demonstrate that the proposed soft EA feet enable controllable attachment of a lightweight micro aerial robot on smooth flat and curved plastic surfaces. At the same time, the results show that substrate properties strongly affect perching performance. In particular, rougher surfaces such as wood and more challenging curved geometries remain limited by the relatively small normal adhesion, indicating that further improvement in normal-force generation will be important in future work.
\section{Conclusions}
This paper presented soft electroadhesive feet for a micro aerial robot perching on smooth and curved surfaces. A simple fabrication process combining laser cutting and film casting was developed to produce soft EA pads with wave-pattern and concentric-circular electrodes in multiple sizes. Pad-level experiments showed that both electrode geometry and pad size influenced the adhesion performance, and that shear force was consistently much higher than normal force under the tested conditions. Among the fabricated designs, the concentric-circular pad Circle~1 provided the best overall balance between force performance and suitability for integration on the Crazyflie platform.

Based on the pad-level results, four Circle~1 pads were integrated onto the landing legs of a Crazyflie 2.1 quadrotor to form soft electroadhesive feet. System-level validation demonstrated controllable attachment on flat plastic surfaces and also confirmed the ability of the soft pads to conform to curved plastic substrates. The robot remained attached as the substrate was rotated to vertical and fully inverted configurations, and detached immediately once the applied voltage was removed. Additional tests also showed that surface properties had a strong influence on performance, with much weaker attachment on wood and reduced stability on more challenging curved surfaces.

Overall, the results highlight the potential of soft electroadhesive feet as a lightweight and compliant solution for MAV perching. Future work will focus on improving normal-force generation, further optimising pad geometry and dielectric design, and reducing the dependence on highly smooth substrates to achieve more reliable attachment in practical environments.
\bibliographystyle{IEEEtran}
\bibliography{Bibliography}
\end{document}